\documentclass[10pt, journal,twocolumn]{IEEEtran}
\usepackage{graphicx}
\usepackage{amsmath}
\usepackage{amssymb}
\usepackage{authblk}
\usepackage{cite}
\usepackage{blindtext}
\usepackage{algorithm}
\usepackage{algpseudocode}
\usepackage{amsmath}
\usepackage{subcaption}
\usepackage{float} 
\usepackage[utf8]{inputenc}

\usepackage{xcolor}
\usepackage{relsize}
\title{ \LARGE  
Cooperative Control of Multi-Quadrotors for Transporting Cable-Suspended Payloads: Obstacle-Aware Planning and Event-Based Nonlinear Model Predictive Control
}
\author{Tohid Kargar Tasooji, IEEE Member, Sakineh Khodadadi, 
	Guangjun Liu, \emph{Senior Member, IEEE}, 
Richard Wang, \emph{Senior Member, IEEE}

\thanks{This work was supported by the Natural Sciences and Engineering Research Council (NSERC) of Canada.
Tohid Kargar Tasooji is with the Department of Aerospace Engineering, Toronto Metropolitan University, Toronto, ON M5B 2K3, Canada (e-mail: tohid.kargartasooji@torontomu.ca).
Sakineh Khodadadi is with the Department of Electrical and Computer
Engineering, University of Alberta, Edmonton,AB,  AB T6G 1H9, Canada (email: sakineh@ualberta.ca).
Guangjun Liu is with the Department of Aerospace Engineering, Toronto Metropolitan University, Toronto, ON M5B 2K3, Canada (e-mail: gjliu@torontomu.ca).
Guanghui Wang is with the Department of Computer Science, Toronto
Metropolitan University, Toronto, ON M5B 2K3, Canada (e-mail: wangcs@torontomu.ca).

 }
}

\begin{document}
\maketitle 
\begin{abstract}
This paper introduces a novel methodology for the cooperative control of multiple quadrotors transporting cable-suspended payloads, emphasizing obstacle-aware planning and event-based Nonlinear Model Predictive Control (NMPC). Our approach integrates trajectory planning with real-time control through a combination of the A* algorithm for global path planning and NMPC for local control, enhancing trajectory adaptability and obstacle avoidance. We propose an advanced event-triggered control system that updates based on events identified through dynamically generated environmental maps. These maps are constructed using a dual-camera setup, which includes multi-camera systems for static obstacle detection and event cameras for high-resolution, low-latency detection of dynamic obstacles. This design is crucial for addressing fast-moving and transient obstacles that conventional cameras may overlook, particularly in environments with rapid motion and variable lighting conditions. When new obstacles are detected, the A* algorithm recalculates waypoints based on the updated map, ensuring safe and efficient navigation. This real-time obstacle detection and map updating integration allows the system to adaptively respond to environmental changes, markedly improving safety and navigation efficiency. The system employs SLAM and object detection techniques utilizing data from multi-cameras, event cameras, and IMUs for accurate localization and comprehensive environmental mapping. The NMPC framework adeptly manages the complex dynamics of multiple quadrotors and suspended payloads, incorporating safety constraints to maintain dynamic feasibility and stability. Extensive simulations validate the proposed approach, demonstrating significant enhancements in energy efficiency, computational resource management, and responsiveness.

\end{abstract}


\section{INTRODUCTION}
\IEEEPARstart{L}ow-cost autonomous micro aerial vehicles (MAVs) equipped with manipulation mechanisms have significant potential to assist humans in various complex and hazardous tasks, including construction [1], transportation and delivery [2], and inspection [3]. In construction, MAV teams can collaborate to transport materials from the ground to upper floors, thereby accelerating the building process. Similarly, MAVs can expedite urgent humanitarian missions or emergency medical deliveries in urban areas by bypassing ground traffic during rush hour and utilizing the unobstructed "highway" in the air. These tasks require aerial robots capable of transporting or manipulating objects.

Additionally, a fleet of aerial robots can provide supplies and set up communication networks in areas with unreliable or nonexistent GPS signals. To increase payload capacity, larger aerial vehicles can be deployed or a group of MAVs can collaborate to carry the cargo. While involving more robots increases system complexity, a team of MAVs can enhance mission resilience, especially if one vehicle experiences a malfunction.

Recent research on quadrotors for cable-suspended payload transportation has shown significant advancements. Loianno et al. [4] developed techniques using quadrotors with cameras and IMUs for stable flight and precise pose estimation, employing nonlinear controllers for efficient control and localization. Li et al. [5] introduced a distributed vision-based control system for independent MAV control and payload estimation, demonstrating scalability in real-world environments. Jin et al. [7] proposed a cooperative control framework for UAVs handling load and safety constraints, ensuring precise tracking and obstacle avoidance. Li et al. [8] created a simulator for aerial transportation, featuring innovative collision models and control algorithms, validated through simulations and experiments.

Model Predictive Control (MPC) has emerged as a promising approach to address challenges in cooperative transportation of cable-suspended payloads using multiple quadrotors, due to its capability to handle multi-variable control problems and enforce constraints systematically. For example, Li et al. [6] presented a NMPC method for managing payloads, optimizing real-time computation, and obstacle avoidance. Erunsal et al. [9] compared linear and nonlinear MPC strategies for trajectory tracking, offering a framework for selecting appropriate MPC methods based on specific needs.

Despite significant advancements in cooperative transportation of cable-suspended payloads with multiple quadrotors and their deployment in these fields, several technical challenges remain unresolved.

A primary challenge is the efficient management of energy and computational resources in multi-quadrotor systems, especially those involving MAVs. This complexity arises from the need to solve nonlinear optimization problems considering system dynamics and constraints like obstacle avoidance and actuator limits. The demand for high update rates to ensure stability and responsiveness, coupled with constrained computational resources onboard MAVs, amplifies this challenge. Traditional periodic update approaches exacerbate energy consumption and communication overhead, critical for MAVs with limited battery capacities and communication bandwidth constraints. To address these challenges effectively, there is a growing interest in event-triggered control systems. Unlike conventional time-triggered methods, event-triggered control updates are activated in response to specific events, such as deviations in trajectories or changes in system dynamics. This adaptive approach conserves energy and optimizes computational resources, enhancing the reliability and scalability of cooperative transportation tasks in MAVs and similar energy-constrained systems. In previous work [12], we presented a novel event-triggered distributed NMPC method for multiple quadrotors handling cable-suspended payloads. This approach addresses challenges such as indirect load actuation and complex dynamics, optimizing computational resources for efficient SE(3) trajectory planning [13]-[18], [24].

A second challenge is achieving accurate reference trajectory tracking in real-time while navigating cluttered and dynamic environments. The complexity is compounded by coupled dynamics between quadrotors and suspended payloads, introducing additional degrees of freedom and potential oscillatory behavior. Ensuring dynamic feasibility and maintaining safety constraints are crucial, especially in the presence of obstacles. Traditional approaches often separate planning and control processes, where a planner generates a feasible trajectory and a controller follows it. However, this separation can introduce latency and reduce responsiveness to sudden environmental changes. Soft-constraint-based methods aim to minimize latency by quickly planning trajectories but may fall into local minima when dynamic obstacles are involved, potentially leading to collisions. Kulathunga et al. [10] enhance MAV navigation in unknown environments by optimizing reference trajectories to avoid obstacles, employing a global planner for trajectory refinement and a local planner for control policy computation. Liu et al. [11] present an Integrated Planning and Control (IPC) framework for quadrotor flight in challenging environments, combining the A* algorithm for local path planning with linear MPC for trajectory and control. Using convex polyhedrons for safety, the MPC rapidly computes control actions (2ms-3.5ms), improving responsiveness and disturbance rejection. Currently, there is no research work addressing cooperative transportation of cable-suspended payloads with multiple quadrotors in a manner integrating planning and control frameworks. Existing literature primarily focuses on control strategies for such systems, exploring how to manage quadrotor and payload movements and stability. However, an integrated approach combining trajectory planning and movement control remains unexplored. This gap suggests an area for further research and development in multi-quadrotor cooperative transportation systems.

Our paper addresses the above challenges in cooperative transportation using MAVs by presenting a novel approach that integrates event-triggered distributed NMPC with an advanced planning system. Our contributions are summarized as follows:
\begin{itemize}
\item We introduce an advanced event-triggered control system that updates based on dynamically generated environmental maps. This system leverages a dual-camera setup with multi-camera systems for static obstacle detection and event cameras for high-resolution, low-latency detection of dynamic obstacles. This event-based update mechanism significantly reduces energy consumption and communication overhead, which is crucial for MAVs with limited battery life and bandwidth.

\item Our methodology pioneers the integration of trajectory planning and real-time control for the cooperative transportation of cable-suspended payloads. By combining the A* algorithm for global path planning with NMPC for local control, we ensure real-time, responsive adjustments to trajectories and control policies. This integration minimizes latency and enhances the system's adaptability to dynamic obstacles and environmental changes.

\item Our NMPC framework effectively manages the complex dynamics between multiple quadrotors and suspended payloads. Incorporating safety constraints into the control formulation ensures dynamic feasibility and stability in cluttered environments. This approach mitigates oscillatory behaviors and maintains precise tracking and obstacle avoidance, which is crucial for safe and efficient operations.

\item We enhance our perception system by incorporating SLAM and object detection techniques. Multi-camera setups detect static obstacles, while event cameras detect dynamic obstacles, triggering control updates only when necessary. This system optimizes resource usage and enables accurate localization and comprehensive environmental mapping. The predictive capability of SLAM and object detection refines collision boundaries and improves the accuracy and safety of MAV navigation in dynamic environments.

\item We validate our approach through extensive simulations, demonstrating significant improvements in energy efficiency, computational resource management, and responsiveness compared to existing approaches. These enhancements make our system suitable for practical applications in construction, delivery, and inspection tasks, where reliable and efficient multi-robot cooperation is essential.

\end{itemize}
By addressing these critical challenges, our paper significantly advances the field of autonomous MAV systems. Our integrated approach not only enhances the state-of-the-art in MAV control strategies but also lays a foundation for future research and development in multi-robot systems operating in complex, dynamic environments.
\begin{figure*}[h!]
\captionsetup{justification=centering}
 \centering \includegraphics[width=0.9 \textwidth]{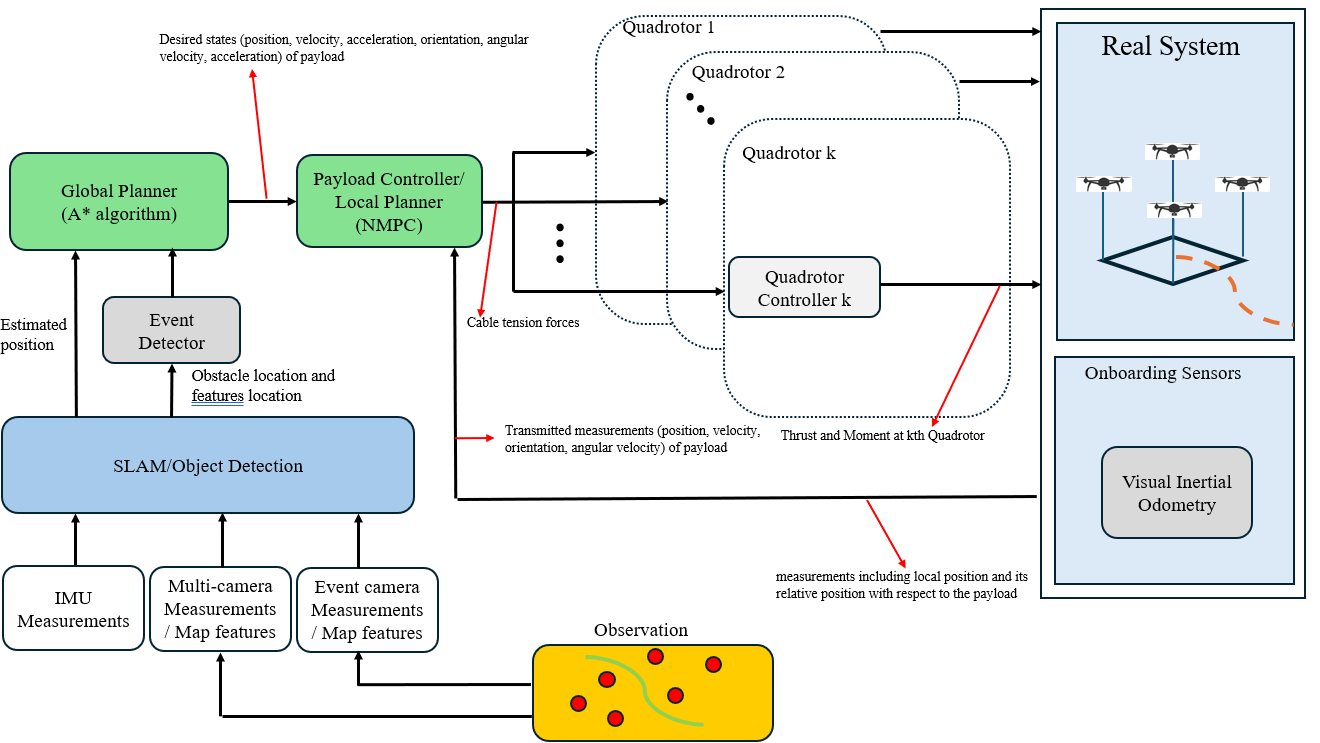}
  \caption{The control block diagram of the proposed approach}  \label{Fig1}
\end{figure*}
\section{PROBLEM FORMULATION}
\subsection{Problem of Interest}
 In this part, we provide an overview of our architecture, which integrates several key components to achieve robust and efficient navigation and control for cooperative transportation tasks involving aerial robots. Our system is tailored for applications such as construction and transportation, and it addresses the gap in integrating trajectory planning with real-time control. The navigation system combines the A* algorithm for global path planning with NMPC for local control, ensuring responsive trajectory adjustments and enhanced obstacle avoidance capabilities. Our system features an event-triggered control mechanism that activates updates based on specific events detected by multi-camera setups and event cameras. This significantly reduces energy consumption and communication overhead, crucial for MAVs with limited battery life and bandwidth. Multi-camera setups detect static obstacles, while event cameras detect dynamic obstacles, triggering control updates only when necessary and optimizing resource usage. Each quadrotor has a multi-camera setup and a sophisticated perception algorithm leveraging Simultaneous Localization and Mapping (SLAM) and object detection. SLAM, based on inputs from multiple cameras, constructs and updates a map of the environment while simultaneously tracking the quadrotors' locations. Object detection refines collision boundaries and improves real-time position estimation, enhancing navigation accuracy in dynamic environments. The Global Planner employs the A* algorithm to generate desired states for the payload, including position, velocity, acceleration, orientation, angular velocity, and acceleration. These desired states guide the overall trajectory of the payload. The Payload Controller/Local Planner operates using NMPC, taking into account the desired states from the Global Planner, transmitted measurements from the payload, and cable tension forces. The NMPC framework produces precise commands for the quadrotor controllers, ensuring optimal local trajectory planning while adhering to the system's dynamic constraints. This involves managing the intricate dynamics of the payload in all six degrees of freedom (6 DoF), and optimizing cable tension forces to achieve the desired thrust and moment commands. Each quadrotor is equipped with its respective Quadrotor Controller, which generates the necessary thrust and moment to execute the planned trajectory and maintain system stability. These controllers are crucial for the fine-tuned manipulation of the payload, ensuring accurate trajectory tracking and system coordination among multiple quadrotors. Onboard sensors provide real-time data, ensuring accurate and up-to-date state estimation. This is vital for maintaining system accuracy and reliability, especially in dynamic and GPS-denied environments where precise navigation is challenging. The real-time data integration allows the system to adapt to changing conditions and maintain robust performance.
 \subsection{System Dynamics}
Consider a team of N (N $\geq$ 3) MAVs collaboratively transporting a rigid payload using cables (see Figure 2). The dynamics are defined as follows \cite{11}:

\begin{equation}
\text{MAVs}
\left\{
\begin{array} {l} 
m_{i}\ddot{p}_{i}(t) = - \text{sat}(F_{i}(t))R(\Theta_{i}(t))e_{z} + m_{i}ge_{z} \\ \ \ \ \ \ \ \ \ \ \ \ \  \ \ \   + T_{i}(t)R(\Theta_{L}(t))e_{i}(t) \\ 
\dot{\Theta}_{i}(t) = \Gamma(\Theta_{i}(t))\omega_{i}(t) \\ 
J_{i}\dot{\omega}_{i}(t) + \mathbb{S}(\omega_{i}(t)) J_{i} \omega_{i}(t) = \tau_{i}(t),
\end{array}
\right.
\end{equation}

\begin{equation}
\text{Load}
\left\{
\begin{array} {l} 
m_{L}\ddot{p}_{L}(t) = m_{L}g e_{z} - \sum_{i=1}^{N} T_{i}(t)R(\Theta_{L}(t))e_{i}(t) \\ 
\dot{\Theta}_{L}(t) = \Gamma(\Theta_{L}(t))\omega_{L}(t)\\
J_{L}\dot{\omega}_{L}(t) + \mathbb{S}(\omega_{L}(t)) J_{L} \omega_{L}(t) = \sum_{i=1}^{N} \mathbb{S}(r_{i})(-T_{i}(t)e_{i}(t))
\end{array}
\right.
\end{equation}

Here, $m_i \in \mathbb{R}^+$ denotes the mass of the $i$th quadrotor $(i = 1, \ldots, N)$, and $J_i \in \mathbb{R}^{3 \times 3}$ is a symmetric positive definite matrix representing inertia. The position and orientation in the inertial reference frame are $p_i(t) = [x_i(t), y_i(t), z_i(t)]^T \in \mathbb{R}^3$ and $\Theta_i(t) = [\phi_i(t), \theta_i(t), \psi_i(t)]^T \in \mathbb{R}^3$, respectively. $R(\Theta_i(t)) \in SO(3)$ is the rotation matrix relating the body-fixed frame to the inertial frame, defined as

\begin{equation}
R(\Theta_i) =
\begin{bmatrix}
c\phi_i c\theta_i & c\phi_i s\theta_i s\psi_i - c\psi_i s\phi_i & c\psi_i c\phi_i s\theta_i + s\phi_i s\psi_i \\
c\theta_i s\phi_i & c\phi_i c\psi_i + s\phi_i s\theta_i s\psi_i & c\psi_i s\phi_i s\theta_i - c\phi_i s\psi_i \\
-s\theta_i & c\theta_i s\psi_i & c\theta_i c\psi_i
\end{bmatrix}
\end{equation}

The angular velocities in the body-fixed frame are $\omega_i(t) = [\omega_{xi}(t), \omega_{yi}(t), \omega_{zi}(t)]^T \in \mathbb{R}^3$, and $\Gamma(\Theta_{i}(t))$ is the transformation matrix linking the body-fixed angular velocity to the derivative of the Euler angles in the inertial frame, given by

\begin{equation}
\Gamma(\Theta_i) =
\begin{bmatrix}
1 & s\phi_i t\theta_i & c\phi_i t\theta_i \\
0 & c\phi_i & -s\phi_i \\
0 & \frac{s\phi_i}{c\theta_i} & \frac{c\phi_i}{c\theta_i}
\end{bmatrix}
\end{equation}

This matrix is well-defined and invertible when $-\pi/2 < \phi_i(t) < \pi/2$ and $-\pi/2 < \theta_i(t) < \pi/2$. Additionally, $g \in \mathbb{R}$ is the gravitational acceleration, and $\mathbf{e}_z = [0, 0, 1]^T \in \mathbb{R}^3$ is the unit vector. $T_i(t) \in \mathbb{R}^+$ is the tension in the $i$th rigid cable, $\text{sat}(a)$ is the saturation function where $a \in \mathbb{R}^+$, and $F_i(t) \in \mathbb{R}^+$ is the thrust of the $i$th quadrotor, subject to the saturation nonlinearity described by

\begin{equation}
\text{sat} (F_i(t)) =
\begin{cases}
F_{\text{max}}, & \text{if } F_i(t) \geq F_{\text{max}} \\
F_i(t), & \text{otherwise}
\end{cases}
\end{equation}

where $F_{\text{max}_i}$ is the maximum thrust limit and $\text{sign}(\cdot)$ is the sign function. Lastly, $\tau_i(t) \in \mathbb{R}^3$ represents the torques of the $i$th quadrotor $(i = 1, \ldots, N)$.

Similarly, $m_L \in \mathbb{R}^+$ is the load mass, and $J_L \in \mathbb{R}^{3 \times 3}$ is the symmetric positive definite load inertia matrix. The load's position and orientation in the inertial reference frame are $\mathbf{p}_L(t) = [x_L(t), y_L(t), z_L(t)]^T \in \mathbb{R}^3$ and $\mathbf{\Theta}_L(t) = [\phi_L(t), \theta_L(t), \psi_L(t)]^T \in \mathbb{R}^3$, respectively. The load's rotational velocity relative to its body-fixed frame is $\boldsymbol{\omega}_L(t) = [\omega_{xL}(t), \omega_{yL}(t), \omega_{zL}(t)]^T \in \mathbb{R}^3$. Additionally, $\mathbf{r}_i \in \mathbb{R}^3$ denotes the attachment point on the payload by the $i$th link, as shown in Figure 2, and $\mathbf{e}_i(t) \in S^2$ is the unit direction vector from the $i$th MAV's center of mass to the $i$th link attachment point.
\begin{figure}[h!]
\captionsetup{justification=centering}
 \centering \includegraphics[width=0.4 \textwidth]{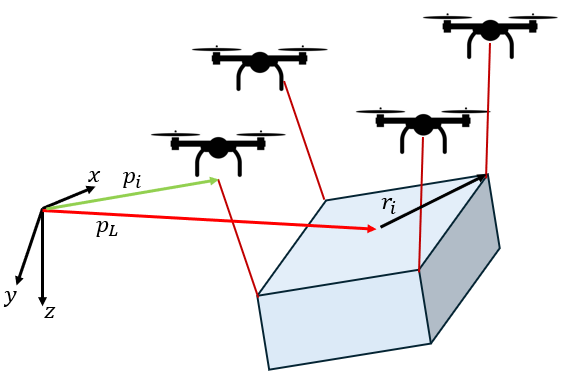}
  \caption{Cable-suspended load transportation by MAVs}  \label{Fig1}
\end{figure} 
\section{Perception System in Cooperative Control of Multi-Quadrotors for Transporting Cable-Suspended Payloads}
The perception system for our cooperative control of multi-quadrotors integrates advanced visual-inertial odometry (VIO) and event-based sensing methodologies to achieve robust obstacle-aware planning and event-based nonlinear model predictive control (NMPC). The system comprises two main components: a multi-camera VIO setup and an event-based camera system, ensuring comprehensive environmental awareness and dynamic obstacle detection.

\subsection{Multi-Camera Visual-Inertial Odometry and SLAM Integration}
The quadrotor is equipped with a multi-camera system designed to enhance visual SLAM capabilities by capturing a wide range of environmental features from multiple perspectives. This setup leverages the methodology from the multi-camera visual SLAM system, addressing the limitations of monocular and stereo visual SLAM systems in complex environments with limited visual features. By integrating multiple cameras pointing in different directions, the system achieves a broader effective field of view (FOV), allowing for more reliable feature detection and robust pose tracking. The cameras are strategically positioned to cover overlapping fields of view, maximizing redundancy and reliability. This redundancy is crucial for maintaining continuous localization and mapping in scenarios where individual cameras might face occlusions or low-feature environments [13].

The visual SLAM system operates with three main threads: tracking, mapping, and back-end processing. The tracking thread handles visual odometry and local map maintenance, interacting closely with the mapping thread, which updates the global map with new keyframes and pose optimizations. The back-end processing thread continuously optimizes the pose graph using the g2o library, ensuring accurate and up-to-date mapping even in dynamic and GPS-denied environments. Additionally, the system incorporates IMUs (Inertial Measurement Units) to complement visual data, providing robust pose estimation and enhancing overall stability.

Key equations and methodologies involved in this system include:

The state of the system is estimated using a combination of visual and inertial measurements. The state vector $\mathbf{x}$ includes the position $\mathbf{p}$, velocity $\mathbf{v}$, and orientation $\mathbf{q}$ of the quadrotor, as well as the biases of the IMU:
\begin{equation}
\mathbf{x} = \left[ \mathbf{p}, \mathbf{v}, \mathbf{q}, \mathbf{b}_g, \mathbf{b}_a \right]
\end{equation}
where $\mathbf{b}_g$ and $\mathbf{b}_a$ are the gyroscope and accelerometer biases, respectively.

The IMU provides high-rate measurements of angular velocity $\boldsymbol{\omega}$ and linear acceleration $\mathbf{a}$. The propagation model updates the state estimate using these measurements:
\begin{equation}
\mathbf{p}_{k+1} = \mathbf{p}_k + \mathbf{v}_k \Delta t + \frac{1}{2} \mathbf{R}(\mathbf{q}_k) (\mathbf{a}_k - \mathbf{b}_a - \mathbf{n}_a) \Delta t^2
\end{equation}
\begin{equation}
\mathbf{v}_{k+1} = \mathbf{v}_k + \mathbf{R}(\mathbf{q}_k) (\mathbf{a}_k - \mathbf{b}_a - \mathbf{n}_a) \Delta t
\end{equation}
\begin{equation}
\mathbf{q}_{k+1} = \mathbf{q}_k \otimes \exp\left(\frac{1}{2} (\boldsymbol{\omega}_k - \mathbf{b}_g - \mathbf{n}_g) \Delta t \right)
\end{equation}
where $\mathbf{R}(\mathbf{q})$ is the rotation matrix corresponding to the quaternion $\mathbf{q}$, $\mathbf{n}_a$ and $\mathbf{n}_g$ are the accelerometer and gyroscope noise, and $\otimes$ denotes quaternion multiplication.

The measurement model relates the 3D point $\mathbf{P}_w$ in the world coordinate system to its projection $\mathbf{p}_c$ in the camera image:
\begin{equation}
\mathbf{p}_c = \mathbf{K} \left[ \mathbf{R}_c \quad \mathbf{t}_c \right] \mathbf{P}_w
\end{equation}
where $\mathbf{K}$ is the camera intrinsic matrix, $\mathbf{R}_c$ and $\mathbf{t}_c$ are the rotation and translation from the world coordinate system to the camera coordinate system.

To handle multiple cameras, features detected in each camera are associated with a common 3D point. The image projection of a map point $ p_j $ to a camera $ C_i $ is:
\begin{equation}
u_{ji} = P_{C_i}(E_{ciw} p_j)
\end{equation}
where $ P_{C_i} $ maps a point in the camera coordinate system $ C_i $ to image coordinates, $ p_j $ are the world coordinates of the map point, and $ E_{ciw} $ is a member of the Lie group SE(3), representing the camera pose in the world coordinate system. The projection error for a feature observed in camera $i$ is minimized:
\begin{equation}
e_{ji} = u_{ji} - \hat{u}_{ji}
\end{equation}
where $\hat{u}_{ji}$ is the predicted feature point in camera $i$.

The optimization problem for the pose update $\mu$ for camera $C_1$ is:
\begin{equation}
\mu_1 = \arg\min_\mu \sum_{i=1}^n \sum_{j \in S_i} \text{Obj} \left( \frac{|e_{ji}|}{\sigma_{ji}}, \sigma_T \right)
\end{equation}
where $\text{Obj}$ is the Tukey biweight objective function, $|e_{ji}|$ is the reprojection error, $\sigma_{ji}$ is the estimated measurement noise, and $\sigma_T$ is a median-based robust standard-deviation estimate of all reprojection errors.

Features are detected using methods such as ORB, FAST, or SIFT, and matched across images using descriptors and RANSAC for geometric verification. The relative pose between frames is estimated using the essential matrix $\mathbf{E}$:
\begin{equation}
\mathbf{E} = \mathbf{K}^\top \mathbf{R} [\mathbf{t}]_\times \mathbf{K}
\end{equation}
where $\mathbf{K}$ is the camera intrinsic matrix, $\mathbf{R}$ is the rotation matrix, and $[\mathbf{t}]_\times$ is the skew-symmetric matrix of the translation vector $\mathbf{t}$.

The back-end processing thread optimizes the pose graph to minimize the reprojection error of feature points. The graph consists of keyframe-pose vertices $V_i$ and relative edges $E_{ij}$ describing the relative pose–pose constraints among vertices. Each vertex $V_i$ stores an absolute pose $E_i$ in the world frame. Pose constraints between keyframes are maintained and optimized. The optimization problem is formulated as:
\begin{equation}
F(E) = \sum_{E \in G_s} \Delta E_{ij}^T \Omega_{ij} \Delta E_{ij}
\end{equation}
where $\Delta E_{ij} := \log(E_{ij} \cdot E_j^{-1} \cdot E_i)$ is the relative pose error in the tangent space of SE(3) and $\Omega_{ij}$ is the information matrix of the pose constraint $E_{ij}$.

The fusion of visual and inertial data is performed using an Extended Kalman Filter (EKF) or an optimization-based approach. The update step for the EKF incorporates visual measurements:
\begin{equation}
\mathbf{K}_k = \mathbf{P}_{k|k-1} \mathbf{H}_k^\top \left( \mathbf{H}_k \mathbf{P}_{k|k-1} \mathbf{H}_k^\top + \mathbf{R}_k \right)^{-1}
\end{equation}
\begin{equation}
\mathbf{x}_{k|k} = \mathbf{x}_{k|k-1} + \mathbf{K}_k \left( \mathbf{z}_k - \mathbf{h}(\mathbf{x}_{k|k-1}) \right)
\end{equation}
\begin{equation}
\mathbf{P}_{k|k} = \left( \mathbf{I} - \mathbf{K}_k \mathbf{H}_k \right) \mathbf{P}_{k|k-1}
\end{equation}
where $\mathbf{K}_k$ is the Kalman gain, $\mathbf{P}_{k|k-1}$ is the predicted error covariance, $\mathbf{H}_k$ is the measurement Jacobian, $\mathbf{R}_k$ is the measurement noise covariance, $\mathbf{z}_k$ is the measurement, and $\mathbf{h}(\mathbf{x})$ is the measurement model.

The relative pose between two camera frames can be estimated using the fundamental matrix $\mathbf{F}$, which relates corresponding points $\mathbf{p}_1$ and $\mathbf{p}_2$ in the two images:
\begin{equation}
\mathbf{p}_2^\top \mathbf{F} \mathbf{p}_1 = 0
\end{equation}
The essential matrix $\mathbf{E}$ can be derived from the fundamental matrix $\mathbf{F}$ if the intrinsic parameters of the cameras are known:
\begin{equation}
\mathbf{E} = \mathbf{K}_2^\top \mathbf{F} \mathbf{K}_1
\end{equation}

The optimization of the bundle adjustment problem is key to refining both the camera poses and the 3D point positions. The objective function is:
\begin{equation}
\min_{\mathbf{T}, \mathbf{P}} \sum_{i=1}^n \sum_{j=1}^m \left\| \mathbf{y}_{ij} - \pi (\mathbf{T}_i, \mathbf{P}_j) \right\|^2
\end{equation}
where $\mathbf{T}_i$ is the pose of the $i$-th camera, $\mathbf{P}_j$ is the $j$-th 3D point, $\mathbf{y}_{ij}$ is the observed 2D point in the $i$-th camera image, and $\pi$ is the projection function.

The integration of inertial measurements into the bundle adjustment framework can further improve the accuracy of the system. This involves augmenting the state vector with IMU states and modifying the cost function to include the IMU residuals.

\subsection{Event-Based Sensing System and SLAM Integration}
Event cameras measure changes in intensity asynchronously, offering high temporal resolution and sparsity, significantly reducing bandwidth and latency. Traditional image-based RGB cameras face a bandwidth-latency trade-off: higher frame rates reduce perceptual latency but increase bandwidth demands. The quadrotor's event-based camera system is pivotal for dynamic obstacle detection and avoidance, leveraging the high temporal resolution and low-latency characteristics of event-based sensors. These cameras capture changes in pixel intensity asynchronously, enabling rapid detection of moving objects and facilitating real-time obstacle avoidance. The event-based camera operates by detecting changes in log intensity at each pixel independently. An event is generated when the change in log intensity exceeds a predefined threshold:
\[
\Delta L(u,t) = \log I(u,t) - \log I(u,t - \Delta t) > C
\]
where $\Delta L$ is the change in log intensity at pixel $u$, $I$ is the intensity, $t$ is the time, and $C$ is the contrast threshold.

Events are represented as a stream of tuples $(x, y, t, p)$, where $(x,y)$ is the pixel location, $t$ is the timestamp, and $p$ is the polarity (indicating whether the intensity increased or decreased). This representation allows for efficient processing of the sparse event data.

\subsection*{System Architecture}

\subsubsection{Event-Based System}

The event-based system for the quadrotor utilizes the high temporal resolution of event cameras for dynamic obstacle detection and avoidance.

\subsubsection{Graph Neural Network (GNN) for Event Processing}

\begin{itemize}
    \item Constructs spatio-temporal graphs from the stream of events.
    \item Uses graph convolutional layers to process the event data.
    \item Applies targeted skipping to prioritize important events and reduce computational load.
\end{itemize}

\subsubsection{Graph Convolution Layers}

The GNN utilizes specialized graph convolutional layers to process event data:
\[
h_i(l+1) = \sigma \left( \sum_{j \in N(i)} W(l) h_j(l) + b(l) \right)
\]
where $h_i(l)$ is the feature vector of node $i$ at layer $l$, $N(i)$ is the neighborhood of node $i$, $W(l)$ is the weight matrix, $b(l)$ is the bias, and $\sigma$ is the activation function.

\subsubsection{Recursive Update Rule}

The asynchronous GNN updates its graph structure and activations recursively, processing each new event individually to reduce computation:
\[
h_i(t+1) = \sigma \left( W h_i(t) + \sum_{j \in N(i)} W_e e_{ij} + b \right)
\]
where $h_i(t+1)$ is the updated feature vector of node $i$ at time $t+1$, $W$ is the weight matrix, $W_e$ is the edge feature between nodes $i$ and $j$, and $b$ is the bias.

\subsubsection{Dynamic Obstacle Detection}

For dynamic obstacle detection, events are clustered and tracked over time to identify and predict the trajectories of moving objects. A probabilistic filtering approach, such as a Kalman filter or particle filter, is employed to estimate the state of each detected obstacle, including its position and velocity.

\textbf{Prediction Step:}
\[
\begin{aligned}
x_{k+1|k} &= A x_k + B u_k + w_k \\
P_{k+1|k} &= A P_k A^T + Q
\end{aligned}
\]

\textbf{Update Step:}
\[
\begin{aligned}
K_k &= P_{k|k-1} H^T (H P_{k|k-1} H^T + R)^{-1} \\
x_k &= x_{k|k-1} + K_k (z_k - H x_{k|k-1}) \\
P_k &= (I - K_k H) P_{k|k-1}
\end{aligned}
\]

where $x_{k+1|k}$ is the predicted state, $P_{k+1|k}$ is the predicted covariance, $A$ is the state transition matrix, $B$ is the control input matrix, $u_k$ is the control input, $Q$ is the process noise covariance, $K_k$ is the Kalman gain, $P_{k|k-1}$ is the prior covariance, $H$ is the measurement matrix, $R$ is the measurement noise covariance, $z_k$ is the measurement, and $I$ is the identity matrix.

The estimated trajectories are used to generate collision-free trajectories for the quadrotors using event-based nonlinear model predictive control (NMPC).

This event-based system for quadrotors efficiently processes dynamic obstacle data in real-time, utilizing event cameras and asynchronous GNNs to enhance performance in complex environments.

\subsection{Global Path Planning with A* Algorithm}

\subsubsection*{Algorithm Overview}
The A* algorithm is used to find the shortest path from the start to the goal in a weighted grid/graph. Nodes represent potential waypoints, and edges represent travel costs.

\subsubsection*{Cost Function}
The total cost \( f(n) \) for a node \( n \) is:
\[
f(n) = g(n) + h(n)
\]
where:
\begin{itemize}
    \item \( g(n) \): Actual cost from the start node to node \( n \).
    \[
    g(n) = g(\text{parent}(n)) + \text{cost}(\text{parent}(n), n)
    \]
    \item \( h(n) \): Heuristic estimate of the cost from node \( n \) to the goal node, usually the Euclidean distance.
    \[
    h(n) = \sqrt{(x_n - x_g)^2 + (y_n - y_g)^2 + (z_n - z_g)^2}
    \]
\end{itemize}

\subsubsection*{Path Generation}
A sequence of waypoints \( \{w_0, w_1, w_2, \ldots, w_G\} \) is generated, guiding the quadrotors from the start node \( w_0 \) to the goal node \( w_G \).

\subsubsection*{Safety Distance Consideration}
To ensure safety, a minimum allowable distance \( r \) between the system and obstacles is incorporated. Cost function adjustments:
\[
\text{cost} = 
\begin{cases} 
\infty & \text{if node } n \text{ is within safety distance } r    \\
\text{standard cost} & \text{otherwise}
\end{cases}
\]

\subsection*{2. State Generation}

\subsubsection*{Position Interpolation}
Cubic spline interpolation is used to create smooth trajectories between waypoints. Desired position \( p_d(t) \) at time \( t \):
\[
p_d(t) = \text{CubicSpline}(\{w_i\}, t)
\]

\subsubsection*{Velocity and Acceleration}
Derived from the interpolated position \( p_d(t) \):
\[
v_d(t) = \frac{dp_d(t)}{dt}
\]
\[
a_d(t) = \frac{d^2p_d(t)}{dt^2}
\]

\subsubsection*{Orientation, Angular Velocity, and Angular Acceleration}
Calculated based on the kinematic and dynamic models of the quadrotors and the payload, ensuring balance and stable flight.

\subsection*{3. Perception and SLAM Integration}

\subsubsection*{Environmental Map Update}
The map \( M(t) \) is continuously updated using SLAM (Simultaneous Localization and Mapping) based on the positions of the quadrotors \( p(t) \) and detected obstacles \( o(t) \):
\[
M(t) = \text{SLAM}(p(t), o(t))
\]
\begin{itemize}
    \item \textbf{SLAM Process:} SLAM combines sensor data from multi-cameras and event-cameras to simultaneously build a map of the environment and localize the quadrotors within this map.
    \item \textbf{Sensor Data:} The perception system integrates data from multi-cameras, event-camera, and IMUs (Inertial Measurement Units) mounted on the quadrotors to gather comprehensive information about the environment.
    \item \textbf{Position Estimation:} The positions of the quadrotors \( p(t) \) are estimated using visual-inertial odometry, which fuses data from multi-cameras and IMUs to provide accurate localization.
    \item \textbf{Map Construction:} Detected obstacles \( o(t) \) are incorporated into the map. The SLAM algorithm refines the map \( M(t) \) based on multi-camera and event-camera data, ensuring it accurately reflects the current state of the environment.
\end{itemize}

\subsubsection*{Obstacle Detection}
Obstacles are detected using the perception system, with the integration of multi-camera and event-camera data to enhance detection accuracy. Their positions \( o_j = (x_j, y_j, z_j) \) are added to the map:
\begin{itemize}
    \item \textbf{Obstacle Identification:} The perception system leverages multi-camera and event-camera data for enhanced obstacle detection. Techniques such as image segmentation, object detection, and 3D point cloud processing are applied to identify obstacles within the environment.
    \item \textbf{Position Estimation:} Once an obstacle is detected, its position \( o_j = (x_j, y_j, z_j) \) is estimated relative to the quadrotors' positions using data from the multi-cameras and event-cameras.
    \item \textbf{Map Update:} The positions of detected obstacles are added to the environmental map \( M(t) \). This allows the system to maintain an up-to-date representation of the environment, including both static and dynamic obstacles.
    \item \textbf{Continuous Monitoring:} The perception system continuously monitors the environment using multi-camera and event-camera data to ensure that new obstacles are detected and added to the map promptly.
\end{itemize}

\subsubsection*{Safety Distance Adjustment}
Integrated into the map, marking regions within the safety distance \( r \) of detected obstacles as high-cost regions to avoid.

\subsection*{4. Event-Triggered Control Updates}

\subsubsection*{Event Detection}
An event \( e(t) \) is triggered when a new obstacle or significant change is detected:
\[
e(t) = \{o_j \mid \text{new obstacle detected at time } t\}
\]

\subsubsection*{Formulation of Event-Detector}
The event-detector is responsible for identifying significant changes in the environment that warrant a recalculation of the path. This is typically done by monitoring the output of the perception system for any newly detected obstacles. The event \( e(t) \) is defined as:
\[
e(t) = \{o_j \mid \text{new obstacle detected at time } t\}
\]
where:
\begin{itemize}
    \item \( o_j \) represents a detected obstacle, characterized by its position \( (x_j, y_j, z_j) \) and possibly other attributes such as size, velocity, etc.
    \item \( t \) is the current time at which the detection occurs.
\end{itemize}

The process of event detection involves:
\begin{enumerate}
    \item \textbf{Continuous Monitoring:} The perception system continuously scans the environment for obstacles using multi-camera and event-cameras
    \item \textbf{Obstacle Identification:} New obstacles are identified by comparing the current sensor data with the previously known map \( M(t-1) \).
    \item \textbf{Event Generation:} If a new obstacle \( o_j \) is detected that was not present in the previous map, it is added to the event set \( e(t) \).
    \item \textbf{Significant Change Detection:} The event set \( e(t) \) is checked. If \( e(t) \neq \emptyset \), it indicates a significant change in the environment that requires a response.
\end{enumerate}

\subsubsection*{Recalculation of Waypoints}
Upon detecting an event, the A* algorithm recalculates the waypoints based on the updated map \( M(t) \):
\[
\{w_1', w_2', \ldots, w_G'\} = \text{A*}(M(t))
\]
\begin{itemize}
    \item The updated map \( M(t) \) includes the newly detected obstacles and changes.
    \item The A* algorithm is re-executed on the updated map to find a new optimal path from the current position to the goal.
    \item The new set of waypoints \( \{w_1', w_2', \ldots, w_G'\} \) is generated, ensuring that the path avoids the new obstacles and adheres to the safety distance \( r \).
\end{itemize}

\subsubsection*{Safety Distance Consideration in Recalculation}
Ensures the new path maintains a safe distance from newly detected obstacles.

The system employs the A* algorithm for optimal path planning, derives desired states from interpolated waypoints, and continuously updates the environmental map using SLAM. An event-triggered mechanism dynamically adapts the path in response to detected changes, ensuring effective navigation, obstacle avoidance, and maintenance of a safety distance around obstacles.

\subsection{Integration and Control}
The perception data from both quadrotors are integrated into a centralized control system employing NMPC for cooperative payload transportation. The NMPC algorithm utilizes the obstacle-aware planning capabilities provided by the multi-camera VIO system and the dynamic obstacle detection from the event-based camera system to generate optimal control inputs. This ensures smooth and collision-free navigation while maintaining the stability and coordination required for transporting cable-suspended payloads. The NMPC framework continuously updates the quadrotors' trajectories based on real-time perception data, considering detected obstacles and the coupled dynamics of the suspended payload. This dynamic adjustment ensures all movements are feasible and the system adheres to safety constraints, optimizing the cooperative transportation process. Onboard sensors continuously provide data for visual-inertial odometry, ensuring accurate state estimation and enabling the system to adapt to changing conditions. The event-triggered mechanism optimizes computational resources by activating control updates based on specific conditions, reducing unnecessary computations and enhancing efficiency. This integrated approach ensures precise trajectory tracking, obstacle avoidance, and robust performance, making it suitable for practical applications in construction, delivery, and inspection tasks. The control system leverages high-precision localization data from the multi-camera VIO system and rapid dynamic obstacle detection from the event-based camera to plan and execute safe and efficient paths. The NMPC algorithm considers the quadrotors' kinematic and dynamic constraints, ensuring all control actions are physically feasible and the payload remains stable throughout the operation. The system's predictive capabilities allow it to anticipate and react to environmental changes promptly, optimizing the cooperative transportation process and minimizing the risk of collisions or payload instability. The control architecture also incorporates a decentralized communication protocol that allows the quadrotors to share perception data and control commands in real-time, ensuring coordinated maneuvers and efficient task execution. The use of decentralized control enhances the system's scalability and robustness, allowing for the addition of more quadrotors without significant modifications to the existing setup.

\subsection*{Payload Nonlinear Model Predictive Control}

We introduce a novel NMPC method for controlling payload pose with quadrotors. NMPC computes state sequence $\{X_0, X_1, \ldots, X_N\}$ and input sequence $\{U_0, U_1, \ldots, U_{N-1}\}$ over a prediction horizon $N$, optimizing a cost function while respecting nonlinear constraints and dynamics [19]-[23]:
\begin{equation}
\min_{X_0,\dots,X_N, U_0,\dots,U_{N-1}} \mathlarger{\mathlarger{\sum}}_{i=0}^{N-1} h(X_i,U_i) + h_N(X_N),\\
\end{equation}
subject to the constraints:
\begin{equation*}
X_{i+1} = f(X_i,U_i), \quad \forall i = 0, \dots, N - 1 \\ 
\end{equation*}
\begin{equation*}
X_0 = X(t_0),\\
\end{equation*}
\begin{equation*}
g(X_i,U_i) \leq 0,
\end{equation*}
where $f(X_i,U_i)$ denotes system dynamics and $g(X_i,U_i)$ represents state and input constraints.

For payload transport with quadrotors, the state vector $X$ and input vector $U$ are defined as:
\[
X = [p_L, \Theta_L, v_L, \omega_L]^T, \quad U = [F, M]^T
\]
where $p_L$ is the position, $\Theta_L$ is the orientation, $v_L$ is the velocity, and $\omega_L$ is the angular velocity of the payload. $F$ and $M$ represent the thrust and moments applied by the quadrotor.

The objective function aims to minimize:
\begin{equation}
\min_{X_i,U_i} e^T _{X_N} Q _{X_N} e_{X_N} + \mathlarger{\mathlarger{\sum}}_{i=0}^{N-1} e^T _{X_i} Q _{X_i} e_{X_i} + e^T _{U_i} Q_{U} e_{U_i}
\end{equation}
where $e_{X_i}$ and $e_{U_i}$ are the state and input errors, respectively:
\begin{equation}
e_{X_i} = \begin{pmatrix} p_{L,des} - p_L \\ v_{L,des} - v_L \\ \log(\Theta_L \otimes \Theta_{des,L}^{-1}) \\ \omega_{L,des} - \omega_L \end{pmatrix}_t^i , \quad e_{U_i} = \begin{pmatrix} F_{des} - F \\ M_{des} - M  \end{pmatrix}_t^i
\end{equation}

\begin{algorithm}
\small
\caption{Proposed Algorithm for the cooperative control of multiple quadrotors transporting cable-suspended payloads, considering obstacle-aware planning and event-based Nonlinear Model Predictive Control }
\begin{algorithmic}[1]
\State \textbf{Initialize} variables and parameters
\While{not goal\_reached}
    \State \textbf{Step 1: Perception and SLAM Integration}
    \State $M(t) = SLAM(p(t), o(t))$
    \State $p(t) = \text{estimate\_position}(\text{multi\_camera\_data}, \text{IMU\_data})$
    \State $o(t) = \text{detect\_obstacles}(\text{multi\_camera\_data}, \text{event\_camera\_data})$
    \State update\_map($M(t), o(t)$)
    
    \State \textbf{Step 2: Global Path Planning with A* Algorithm}
    \State waypoints = \text{A\_star\_algorithm}($M(t), \text{start\_node}, \text{goal\_node}$)
    \State path = \text{cubic\_spline\_interpolation}(waypoints)
    \State velocities, accelerations = \text{derive\_velocity\_acceleration}(path)
    
    \State \textbf{Step 3: Dynamic Obstacle Detection}
    \For{each event in \text{event\_camera\_data}}
        \If{new\_obstacle\_detected(event)}
            \State update\_map($M(t), \text{event}$)
            \State trigger\_event($e(t)$)
        \EndIf
    \EndFor
    
    \State \textbf{Step 4: Event-Triggered Control Updates}
    \If{event\_detected}
        \State waypoints = \text{recalculate\_waypoints}($M(t)$)
        \State path = \text{cubic\_spline\_interpolation}(waypoints)
    \EndIf
    
    \State \textbf{Step 5: Payload Nonlinear Model Predictive Control (NMPC)}
    \For{$i = 1$ to \text{prediction\_horizon}}
        \State $X_i, U_i = \text{NMPC\_optimize\_cost\_function}()$
        \State apply\_control\_inputs($U_i$)
    \EndFor
    
    \State \textbf{Step 6: Quadrotor Control}
    \For{each quadrotor}
        \State compute\_desired\_tension\_forces()
        \State apply\_control\_actions()
    \EndFor
\EndWhile
\State \textbf{Finalize}
\State finalize\_trajectory()
\end{algorithmic}
\end{algorithm}

\subsection*{Quadrotor Control}

After solving NMPC, quadrotors compute states $\{X_0, X_1, \ldots, X_N\}$ and inputs $\{U_0, U_1, \ldots, U_{N-1}\}$ over $N$ steps. Inputs $U_0$ define desired cable tension forces $\mu_{\text{des}_k}$ per quadrotor cable:
\begin{equation}
\begin{array}{l}
\mu_{k,\text{des}} = \text{diag}(R_L, \ldots, R_L) P^\dagger \begin{pmatrix} R_{L}^T F_0 \\ M_0 \end{pmatrix}
\end{array}
\end{equation}
where $R_L$ is the rotation matrix, and $P^\dagger$ is the pseudo-inverse of $P$.

Tension forces $\mu_k$ and desired directions $\xi_{k,\text{des}}$ and angular velocities $\omega_{k,\text{des}}$ are computed for each cable link:
\begin{equation}
\begin{array}{l}
\mu_k = \xi_k \xi_k^T \mu_{k,\text{des}}
\end{array}
\end{equation}
\begin{equation}
\begin{array}{l}
\xi_{k,\text{des}} = -\frac{\mu_{k,\text{des}}}{\|\mu_{k,\text{des}}\|}, \quad \omega_{k,\text{des}} = \xi_{k,\text{des}} \times \dot{\xi}_{k,\text{des}}
\end{array}
\end{equation}
Quadrotor thrust $f_k$ and moments $M_k$ are computed as:
\begin{equation}
\begin{array}{l}
f_k = u_k \cdot R_ke_3
\end{array}
\end{equation}
\begin{equation}
\begin{array}{l}
M_k = K_Re_{R_k} + K_\Omega e_{\Omega_k} + \Omega_k \times J_k\Omega_k \\ - J_k(\hat{\Omega}_kR_k^T R_{k,\text{des}}\Omega_{k,\text{des}}   - R_k^T R_{k,\text{des}}\dot{\Omega}_{k,\text{des}})
\end{array}
\end{equation}
where $u_k$ is the control input, $R_k$ is the rotation matrix, $e_3$ is the unit vector in the $z$-direction, $K_R$, $K_\Omega$, $e_{R_k}$, and $e_{\Omega_k}$ are control gains and errors.

\section{Simulation Results}
The simulation models the behavior of four quadrotors navigating a complex three-dimensional (3D) environment while transporting a payload. Each quadrotor starts from one of four distinct positions within a $100 \times 100 \times 100$ unit grid: (0, 0, 0), (0, 10, 0), (10, 0, 0), and (10, 10, 0), aiming to converge at a common destination at (90, 90, 90). Each quadrotor has a maximum acceleration parameter of 1.0 units per second squared (m/s²) and utilizes a Nonlinear Model Predictive Control (NMPC) algorithm to compute its position based on real-time control inputs.

The payload, weighing 232 grams, is suspended from the quadrotors by 1-meter cables. Given that the payload's mass exceeds the capacity of any individual quadrotor, they collaboratively support and stabilize it throughout the flight. The dynamics of the payload are coupled with those of the quadrotors, with the forces transmitted through the cables ensuring coordinated movement and stability.

The simulation environment features multiple static obstacles, including cubic regions extending from (20, 20, 20) to (24, 24, 24) and from (30, 40, 40) to (34, 44, 44). Additionally, six dynamic obstacles are present, each following specific movement patterns. For example, Dynamic Obstacle 1 begins at (10, 10, 10) with a cyclical movement pattern, while Dynamic Obstacle 2 starts at (90, 90, 90) and follows a similar cyclical trajectory. Dynamic Obstacle 3 exhibits a diagonal movement from (50, 30, 30). The presence of these dynamic obstacles adds complexity to the navigation task, necessitating advanced path-planning strategies.

Path planning is achieved using the A* algorithm, which computes trajectories from the initial positions of the quadrotors to the goal, accounting for both static and dynamic obstacles. The innovation in this work lies in the integration of dynamic waypoint generation with real-time obstacle updates. The A* algorithm not only determines initial trajectories but also adapts them in response to the movement of dynamic obstacles and changing environmental conditions. This dynamic waypoint generation enables the quadrotors to adjust their paths effectively as new obstacles are detected or existing ones move.

The control framework employs the NMPC algorithm to compute real-time control inputs for each quadrotor. This framework manages the quadrotors' trajectories and ensures the coordinated transport of the payload. The NMPC algorithm facilitates adherence to planned paths and allows for dynamic adjustments in response to unforeseen obstacles. Additionally, the simulation integrates a Simultaneous Localization and Mapping (SLAM) module, which continuously updates the environmental map to maintain accurate navigation. While SLAM is crucial for operational accuracy, the primary focus is on validating the effectiveness of the proposed NMPC-based control methods.

The simulation results highlight the effectiveness of the proposed multi-quadrotor control framework:

\begin{itemize}
    \item \textbf{Trajectory Tracking and Collision Avoidance:} The quadrotors successfully followed their computed trajectories, effectively avoiding collisions with both static and dynamic obstacles. This was verified by comparing the planned paths with the actual trajectories, demonstrating minimal deviations and successful obstacle avoidance.
    
    \item \textbf{Dynamic Path Adjustment:} Upon detecting dynamic obstacles, the event detector generated new waypoints, enabling the quadrotors to dynamically adjust their paths in real-time. Results show that the quadrotors effectively re-routed their trajectories to avoid moving obstacles, significantly improving obstacle avoidance and path adaptability.
    
    \item \textbf{Performance Metrics:} The effectiveness of the NMPC algorithm and dynamic waypoint generation was quantified by measuring the quadrotors' success in reaching the goal while maintaining safe distances from obstacles. The results indicated a marked improvement in navigation efficiency, with robust performance in managing complex and dynamically changing environments.
\end{itemize}

These results underscore the effectiveness of the proposed control framework and path-planning methods, demonstrating their capability to address real-world navigation challenges and enhance decision-making processes in dynamic environments.

\begin{figure}[h]
\captionsetup{justification=centering}
 \centering \includegraphics[width=0.4 \textwidth]{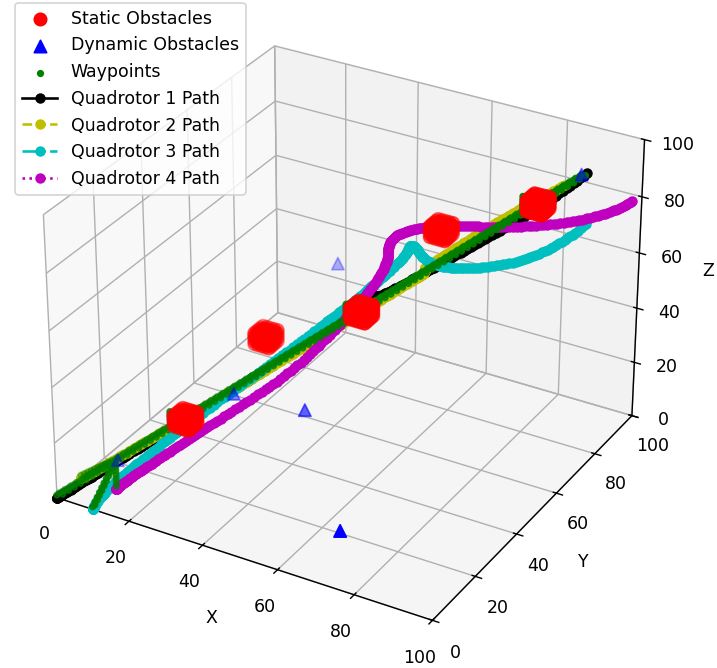}
  \caption{Event-triggered nonlinear model predictive control with proposed obstacle-aware planning for position trajectory of quadrotors in 3D in the presence of static and dynamics obstacles }  \label{Fig1}
\end{figure} 

\section{Conclusion}
In this paper, we presented a comprehensive methodology for the cooperative control of multiple quadrotors transporting cable-suspended payloads in dynamic and cluttered environments. Our approach integrates advanced trajectory planning with real-time control, leveraging a combination of the A* algorithm for global path planning and Nonlinear Model Predictive Control (NMPC) for local control. This integration enhances the system's adaptability, ensuring real-time adjustments to trajectories and control policies in response to environmental changes, particularly the detection of both static and dynamic obstacles.

The proposed system features an event-triggered control mechanism that updates based on specific events identified through dynamically generated environmental maps. These maps, constructed using a dual-camera setup, enable efficient detection of obstacles, optimizing energy consumption and computational resources. The integration of Simultaneous Localization and Mapping (SLAM) and object detection techniques within this framework allows for precise localization and comprehensive environmental mapping, further improving the accuracy and safety of the system.

Extensive simulations validate the effectiveness of our approach, demonstrating significant improvements in energy efficiency, computational resource management, and responsiveness compared to existing methodologies. The results confirm that our integrated framework effectively addresses the complex dynamics of multiple quadrotors and suspended payloads, maintaining stability and safety constraints even in challenging environments.

 The proposed approach not only enhances the reliability and efficiency of multi-robot cooperation but also provides a robust foundation for future research and development in multi-robot systems operating in complex, dynamic environments.

\end{document}